\documentclass[conference]{IEEEtran}
\IEEEoverridecommandlockouts

\usepackage{cite}
\usepackage{graphicx}
\usepackage{amsmath,amssymb,amsfonts}
\usepackage{algorithmic}
\usepackage{textcomp}
\usepackage{xcolor}
\usepackage{caption}
\usepackage{multirow}
\usepackage{booktabs}
\usepackage{mathtools}

\begin{document}

\title{Forecasting Unseen Points of Interest Visits Using Context and Proximity Priors\\
\thanks{This research project has benefited from the Microsoft Accelerate Foundation Models Research (AFMR) grant program through which leading foundation models hosted by Microsoft Azure along with access to Azure credits were provided to conduct the research.}}

\author{
    Ziyao Li\\
    \textit{Westridge School}\\
    Pasadena, USA\\
    emilyli1072@gmail.com
\and
    Shang-Ling Hsu\\
    \textit{University of Southern California}\\
    Los Angeles, USA\\
    hsushang@usc.edu
\and
    Cyrus Shahabi\\
    \textit{University of Southern California}\\
    Los Angeles, USA\\
    shahabi@usc.edu
}

\maketitle

\begin{abstract}
Understanding human mobility behavior is crucial for numerous applications, including crowd management, location-based recommendations, and the estimation of pandemic spread. Machine learning models can predict the Points of Interest (POIs) that individuals are likely to visit in the future by analyzing their historical visit patterns. Previous studies address this problem by learning a POI classifier, where each class corresponds to a POI. However, this limits their applicability to predict a new POI that was not in the training data, such as the opening of new restaurants. To address this challenge, we propose a model designed to predict a new POI outside the training data as long as its context is aligned with the user's interests. Unlike existing approaches that directly predict specific POIs, our model first forecasts the semantic context of potential future POIs, then combines this with a proximity-based prior probability distribution to determine the exact POI. Experimental results on real-world visit data demonstrate that our model outperforms baseline methods that do not account for semantic contexts, achieving a 17\% improvement in accuracy. Notably, as new POIs are introduced over time, our model remains robust, exhibiting a lower decline rate in prediction accuracy compared to existing methods.

\end{abstract}

\begin{IEEEkeywords}
Human mobility modeling, Point-of-Interest forecasting, next visit prediction
\end{IEEEkeywords}

\section{Introduction}

POI forecasting is important for understanding and predicting human mobility patterns, which is valuable across various industries and applications. By forecasting the next POI a person or group of people is likely to visit, businesses can make more informed decisions, improve services, and optimize resources. When used iteratively, the forecasting models can generate a synthetic sequence of visits. These models are critical to crowd management, location-based recommendations, pandemic spread estimation, and traffic flow prediction, to name just a few. Specifically, accurate predictions of future POIs can enhance personalized advertising for various venues. 

Unlike dense trajectory data collected passively from mobile phones or car GPS~\cite{kiukkonen2010towards,zheng2011geolife}, \textit{sequences of visits} are typically used for POI forecasting, where each \textit{visit} represents a user check-in at a specific POI, collected by some location-based social networks (LBSNs) such as Foursquare\cite{foursquare_dataset}. 
Many prior studies approach the task of next POI forecasting by training classifiers, exploring the impact of input features, model architectures, and training techniques on classification performance. More recent work leverages Transformer-based models as a powerful backbone to enhance the model's understanding of visit sequences~\cite{xue2021mobtcast,getnext,deepjmt,starhit}.

However, due to the system's heavy reliance on known training (i.e., seen) data, its performance significantly deteriorates when a new POI emerges. For example, if a new boba shop opens, nearby users who enjoy boba tea may be inclined to visit. In such cases, the model should be able to predict the boba shop as a potential next POI. However, since the model has not been trained on data that includes the new POI, it lacks awareness of its existence. Consequently, the model cannot assign probabilities to these new POIs. Therefore, unless the most up-to-date data, including visits to new POIs, are continuously available and the model is regularly retrained, its ability to accurately forecast future locations will degrade rapidly. In practice, given the limited availability of real-world sequence-of-visit mobility data, these conditions are rarely met.

To overcome this limitation, we propose a novel approach for forecasting the next POI in a sequence of visits by leveraging both the semantic category and location of each POI. Our method follows a simple heuristic: when choosing the next POI, a user is likely to consider both the category and proximity of available POIs. For instance, at noon, the user is more likely to visit a restaurant than a mall, and they would probably prefer a nearby restaurant over a distant one. 
While we follow prior work in using a Transformer-based model~\cite{vaswani2017attention} to encode the social and semantic contexts of a user's past visit sequences, instead of directly predicting the next POI--which can lead to missing a new POI--we design the training objective to classify POI \textit{semantic categories}. We then aggregate the distribution of trip distances as \textit{proximity priors} and combine both in a probabilistic framework to predict the next POI.

To evaluate the effectiveness of our proposed method, we simulate new POIs by (1) varying a temporal threshold in the POI dataset, (2) using that threshold to split the train/test sets, and (3) defining a POI as ``new'' if it only appears after the threshold (i.e., it does not appear in the training set). Since our model incorporates both the semantic context and proximity information of POIs, it can predict a new POI during the inference phase, even if the POI was unseen during training. This refinement is particularly advantageous for location-based recommendation systems, where continuously updating human trajectory data and models may not be desirable. By enabling our model to recommend a user's next destination, a newly opened POI is not overlooked due to unfamiliarity, thereby better reflecting how humans make decisions.

In summary, the contributions of our work are threefold: (1) we introduce a novel method that combines semantic categories and proximity information to enhance prediction accuracy, (2) we develop a Transformer-based model capable of predicting previously unseen POIs, mitigating the decline in accuracy as more unseen POIs start to populate, and (3) our approach offers a practical solution for location-based recommendation systems by improving the model's ability to recommend newly opened POIs, even in the absence of continuously updated trajectory data. These advancements align the model’s behavior more closely with human decision-making processes, offering significant potential for real-world applications.

\section{Related Work}
\label{sec:related-work}

Researchers have explored various strategies to address the challenge of human mobility modeling. 
Two popular human mobility modeling tasks are trajectory synthesis and next visit forecasting. 

\paragraph{Trajectory Synthesis}
The task of trajectory synthesis involves synthesize a human mobility trajectory, where each point contains a location and a timestamp, which can be as dense as one point per second.~\cite{lin2023generating,long2023practical}
For example, MoveSim~\cite{lin2023generating} proposes a framework that contains a urban structure modeling component in the generator for simulating the mobility behavior and mobility regularity-aware loss in the discriminator for distinguishing generated mobility trajectory from real mobility trajectories. Pretrained mobility strategies are also included to increase the accuracy of predictions. 
Long et al.~\cite{long2023practical} proposed a trajectory generation mechanism based on a variational autoencoder (VAE), namely VOLUNTEER, utilizing variational point processes to generate user trajectories. The model consists of two key components: the User VAE and the Trajectory VAE. The User VAE captures user-level features through a transformer and extracts relevant user attributes. Meanwhile, the Trajectory VAE generates a sequence of points representing location and time, learning the distribution parameters necessary for accurate trajectory prediction. It also accounts for time intervals by modeling dwell time (as a probability distribution) and travel time, while incorporating various travel modes. The model integrates classical temporal point processes with neural networks to enhance its predictive accuracy.

\paragraph{Next Visit Forecasting}
Another line of human mobility modeling research studies the next visit forecasting problem~\cite{Hsu2024TrajGPT,cho2011friendship,feng2018deepmove,liu2016predicting,ying2014mining,getnext,starhit,deepjmt,mobtcast,zhao2020go,deepmove,stan,stgn,asppa,llmmob}. We can roughly define the task of next visit forecasting as follows: Given a sequence of visits, where each visit is associated with a location and a time span, such as staying at an office from 9am to 5pm, forecast the next visit of the sequence.
Some researchers in this field focuses on POI forecasting, which is a special case of this problem, as they only model visits that are associated with a POI.~\cite{cho2011friendship,feng2018deepmove,liu2016predicting,ying2014mining,getnext,starhit,deepjmt,mobtcast,zhao2020go,deepmove,stan,stgn,asppa,llmmob}. 
A notable study is MobTCast~\cite{xue2021mobtcast}, which identifies and integrates several types of contexts for POI prediction. 
They use a Transformer-based feature extractor to obtain the representations of semantic and social mobility features, of which the latter aggregates the visited POI history and semantic category information of other users. 
Then, they define two training objectives: one is POI location regression, and the other is POI classification. The former is an auxilarity task that is performed only during training.
We adapt the visit sequence encoding part of MobTCast to convert the temporal, semantic, social, and spatial contexts in the input sequence of visits into a visit sequence representation.

While prior studies excel in forecasting POIs that exist in the training data, because of the heavy reliance on exposure to POIs during the training phase, many of these methods struggle to accurately predict future visits for POIs unseen in training.
To this end, instead of directly predicting the POI and spatial location, we first determine the semantic context and then probabilistically combine it with pre-computed proximity priors to predict the next POI. In this way, newly opened POIs can be predicted although it was previously unfamiliar to the model. 


\section{Problem Formulation}
First, we define a set of POIs \(\mathbf{P}\) as
\begin{equation}
\mathbf{P} := \left\{ (\mathrm{poi}_k,\ \mathrm{lat}_k,\ \mathrm{lon}_k,\ \mathrm{cat}_k) \mid k \in \{1, \ldots, C_{\mathrm{P}}\} \right\}
\end{equation}
where \( \mathrm{poi}_k \) is the ID of the POI, \( \mathrm{lat}_k \) is the latitude of the POI, \( \mathrm{lon}_k \) is the longitude of the POI, \( \mathrm{cat}_k \) is the category code of the POI,\footnote{We use the POI category codes from the Fouresquare dataset~\cite{foursquare_dataset}.} and \( C_{\mathrm{P}} \) is the total number of POIs.

Then, define the sequence of POI visits made by user $i$ as:
\begin{equation}
    \mathbf{v}_{i} := \left\{ (\mathrm{user}_i,\ \mathrm{time}_j,\ \mathrm{poi}_{j}^{i}) \mid j \in \{1, \ldots, n_i\} \right\}
\end{equation}
where $\mathrm{user}_i$ is the user ID, $\mathrm{time}_i$ is the check-in timestamp of the visit, \( n_i \) is the total number of visits by user $i$.
Building upon $\mathbf{v}_i$, we define an $m$-user sequence-of-visit dataset \( \mathbf{V} \) as:
\begin{equation}
    \mathbf{V} := \left\{ \mathbf{v}_{i}  \mid i \in \{1, \ldots, m\} \right\}
\end{equation}

Lastly, we define the task of next POI forecasting as follows:
Given a set of visits $\mathbf{V}$ and a set of POIs $\mathbf{P}$, predict the POI code of the next visit by each user:\footnote{Assuming $\mathbf{P}$ to be a superset of POIs in $\mathbf{V} \cup \mathbf{V}^{*}$.}
\begin{equation}
    \mathbf{V}^* := \left\{ (\mathrm{user}_i,\ \mathrm{poi}^{i}_{n_i+1} ) \mid i = 1,...,m\right\}
\end{equation}


\section{Background}
\label{sec:background}

In a next POI forecasting task, the solution is typically designed as a POI classifier, which is trained using cross-entropy loss. Cross entropy loss is designed to minimize the discrepancy between the predicted probability distribution and the true distribution of the target classes (POIs).

For each input \( \mathbf{v}_i \in \mathbf{V} \), the model outputs a vector of logits (i.e. raw scores assigned by the model) \( z \in \mathbb{R}^{C_{\mathrm{P}}} \), which are converted to class probabilities \( \hat{p} \) using the softmax function:
\begin{equation}
    \hat{p}_k := \frac{\exp(z_k)}{\sum_{j=1}^{C_{\mathrm{P}}} \exp(z_j)}
\end{equation}
where \( \hat{p}_k \) is the predicted probability of the next visit being to POI \( k \), and \( z_j \) is the logit corresponding to POI \( j \). The model is trained to minimize the cross-entropy loss, defined as:
\begin{equation}
    \mathcal{L} := - \sum_{k=1}^{C_{\mathrm{P}}} y_k \log(\hat{p}_k)
\end{equation}
where \( y_k \in \{0, 1\} \) is the ground-truth indicator for POI \( k \).

When training converges, i.e., when the training loss approaches zero, the model is expected to assign high probabilities to the POIs that are actually associated with the user visits in the training data. However, for classes (POIs) that were \emph{never visited} in the training data, the model has no positive training examples. In this scenario, the probabilities assigned to such classes will tend to approach zero as the model focuses on the POIs present in the training data.\footnote{Assuming a typical model without explicit regularization on unseen classes.}

This is because the absence of positive signals for unseen classes implies that the corresponding logits \( z_i \) for such POIs will likely remain low relative to the logits for visited POIs. After applying the softmax, this results in very low probabilities for these unseen classes. Specifically, as the loss converges, the difference between logits for visited and unvisited POIs increases, driving the probabilities for unvisited POIs closer to zero:
\begin{equation}
    \hat{p}_{\text{unvisited}} := \frac{\exp(z_{\text{unvisited}})}{\sum_{j=1}^{C_{P}} \exp(z_j)} \approx 0
\end{equation}

Hence, when training on user visits to a subset of POIs, the model implicitly learns to ignore POIs that were not part of the training visits. Therefore, the probabilities assigned to these unseen classes approach zero as the training loss converges, reflecting the model's confidence that these POIs are unlikely to be associated with the next visit.

In this study, we propose to classify the \emph{categories} of POIs, rather than individual POIs themselves. This approach is motivated by the observation that the number of POI categories is likely orders of magnitude smaller than the total number of individual POIs~\cite{choi2014poi}. By focusing on categories, we can mitigate the requirement for training-phase exposure to individual POIs in order to predict them, while still capturing meaningful patterns in user behavior across an ever-changing environment.

\section{Method}\label{sec:method}
\begin{figure*}[tb]
    \centering
    \includegraphics[width=0.8\linewidth]{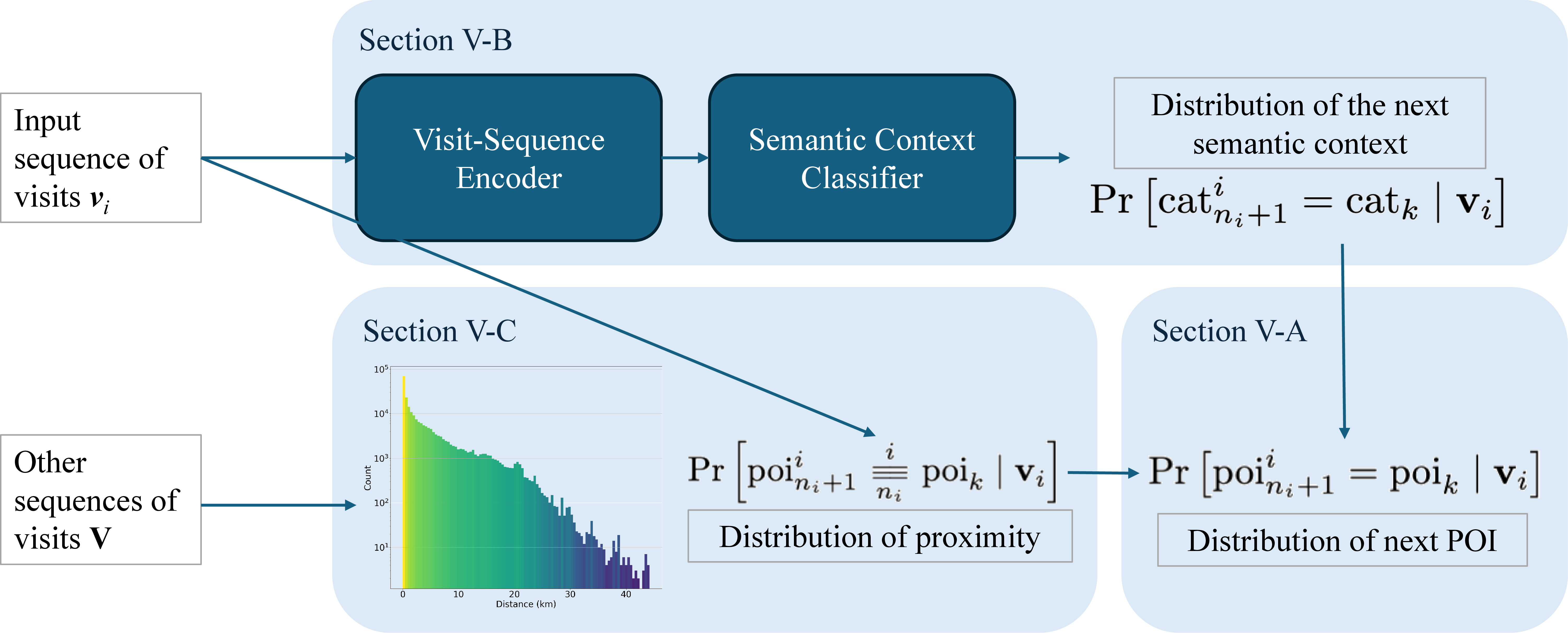}
    \caption{Our proposed model.}
    \label{fig:model}
\end{figure*}
The overall framework of the proposed method is illustrated in Fig.~\ref{fig:model}, which consists of three parts:
(i) \textbf{Joint Probability Approximation} (Section~\ref{sec:combining}): the semantic context probability and the proximity prior probability are combined to predict the next POI.
(ii) \textbf{Semantic Context Prediction} (Section~\ref{sec:predicting-semantic-context}): this is used to generate a probability vector of possible POI categories.
(iii) \textbf{Proximity Prior Computation} (Section~\ref{sec:proximity-priors}): our approximated prior distribution of the distance between the locations of two consecutive visits.
We present each of the above in detail in the following subsections.

\subsection{Joint Probability Approximation}
\label{sec:combining}
\newcommand{\equiiiv}{\ \equiv\joinrel\equiv\ }
\newcommand{\eqdist}[2]{\substack{\mathclap{#1} \\ \equiiiv \\ {#2}}}

To predict the next POI without training a POI classifier, we approximate the POI distribution with (1) the distribution of the next POI categories and (2) the distribution of the distance from the current POI to the next POI.

For simplicity, we first define a binary operator $\eqdist{i}{j}$ that takes two POIs $a,b$ within the Region of Interest (ROI) as input, and outputs whether their distances to the $j$-th location of user $i$ (i.e. the location of the $j$-th visit $\mathrm{poi}^i_{j}$) are similar:
\begin{equation}
    \left[ a \eqdist{i}{j} b \right]
    := \left[ \mathrm{distance}\left(a, \mathrm{poi}^{i}_{j} \right) \simeq \mathrm{distance}\left(b, \mathrm{poi}^{i}_{j}\right) \right]
\end{equation}
where $\mathrm{distance}\left( a, b \right)$ denotes the distance between location $a$ and location $b$.\footnote{We use the Euclidean distance in Universal Transverse Mercator (UTM) coordinate system.}

Then, we make a conditional independence assumption between (1) the distribution of the category of the next POI and (2) that of its distance to the user's last location:
\begin{align}
    \notag &\Pr \left[\mathrm{poi}^{i}_{n_{i}+1} = \mathrm{poi}_{k} 
        \mid \mathbf{v}_i \right] \\
    \approx\ &\Pr \left[ 
        \left( \mathrm{cat}^{i}_{n_{i}+1} = \mathrm{cat}_{k} \right)\wedge
        \left( \mathrm{poi}^{i}_{n_{i}+1} \eqdist{i}{n_{i}} \mathrm{poi}_{k} \right)
        \mid \mathbf{v}_i \right] \\
    \approx\ & \Pr \left[\mathrm{cat}^{i}_{n_{i}+1} = \mathrm{cat}_{k} 
        \mid \mathbf{v}_i \right]
        \Pr \left[ \mathrm{poi}^{i}_{n_{i}+1} \eqdist{i}{n_{i}} \mathrm{poi}_{k} 
        \mid \mathbf{v}_i \right] \label{eq:two-terms}
\end{align}
where $\Pr[\cdot]$ denotes a probability function.

We approximate each of the two terms in (\ref{eq:two-terms}) independently in the following subsections. As shown in the same equation, we then arithmetically combine both terms to predict the distribution of the next POI.



\subsection{Semantic Context Prediction}
\label{sec:predicting-semantic-context}
To model the distribution of the categories (semantic context) of the next POI visit $\Pr \left[\mathrm{cat}^{i}_{n_{i}+1} = \mathrm{cat}_{k} \mid \mathbf{v}_i \right]$, which is the first term in (\ref{eq:two-terms}), we first extract complex social and semantic mobility features of $\mathbf{v}_i$, the input visit sequence of user $i$, and transform them into a vector representation of the sequence of visits. 
Then, we design a semantic context classification objective, predicting the semantic context with a multilayer perceptron (MLP). 
The parameters of the visit sequence encoder and the semantic context classifier are learned in an end-to-end fashion.

\subsubsection{Visit Sequence Encoder}
To encode the sequence of visits, we adopted MobTCast’s semantic-aware mobility feature extractor, which processes three inputs: the POI itself, the timestamp, and the POI's semantic category, and produces a vector representation. We summarize sequence encoding procedure in the following.

Overall, the user’s history trajectory is modeled by a mobility feature extraction process that incorporates not only the visited POIs and timestamps, but the semantic context of the POIs. The key idea is that the semantic similarity between different POIs provides valuable insight for predicting future visits, especially in a dynamic context. For example, one who usually visits a restaurant on Friday evenings might be interested in visiting a newly-opened restaurant on a future Friday evening. Although those restaurants are different POIs, their semantic context can all be labeled as Food.

First, we encode the sequence of visits of a user into a vector representation.
The inputs are embedded into embeddings (POI, category, and temporal embeddings), which are then concatenated. A Transformer Encoder is then employed to encode the user’s sequence-of-visit history by transforming the concatenated embeddings into one vector representation. The self-attention mechanism in the Transformer allows the model to capture dependencies across the entire sequence of visits, and by design, its positional encoding allows itself to capture the temporal order of visits.

Then, we leverage enhance a visit sequence representations with that of the users having similar visit patterns.
To find what users exhibit similar patterns of visits, we construct a co-location matrix and leverage cosine similarity.
The co-location matrix is constructed as a user-POI frequency matrix of shape $m \times C_{\mathrm{P}}$, where the entry at $[i, j],\ i \in \{1, \ldots, m\},\ j \in \{1, \ldots, C_{\mathrm{P}}\}$ contains the number visits user $i$ pays to POI $j$.
We take each row of this matrix as one user vector, and use the cosine similarity of pairs of user vectors to determine their visit-pattern similarity.
To one user, we define the set of other users with high visit-pattern similarity as their ``neighbors.''

In the social context extraction process, we first embed the visit sequence of a user and that of their neighbors. Then, we use a multi-head self-attention mechanism to weigh and fuse the visit representations of their neighbors, producing $f_i$, the social-context-enhanced sequence-of-visit representation of user $i$.

\subsubsection{Semantic Context Classification}  
Given the context-aware feature \( f_{i} \), we forecast the POI semantic context of user $i$ at the next time step $n_{i}+1$ by predicting the probability distribution of the category of the next POI:
\begin{equation}
\hat{\Pr} \left[ \mathrm{cat}^i_{n_{i}+1} \mid \mathbf{v}_i \right] := \text{softmax}(\text{MLP}(f_{i}))
\end{equation}
where $\hat{\Pr}$ denotes the predicted probability distribution.

We treat the next semantic context prediction as a classification over all semantic contexts, so we use multi-class cross entropy loss to obtain the prediction loss for semantic contexts:
\begin{equation}
\mathcal{L} := - \sum_{s=1}^{C_{\mathrm{S}}} y_s \log(\hat{\Pr} \left[ \mathrm{cat}^i_{n_{i}+1} = \mathrm{cat}_s \mid \mathbf{v}_i \right])
\end{equation}
where $C_{\mathrm{S}}$ denotes the number of distinct semantic contexts (POI categories), \(y_s\) represents the ground truth (0 or 1) that the user \( i \) will visit the POI with the \( s \)-th semantic context $\mathrm{cat}_s$ next.

\subsection{Proximity Prior Computation}
\label{sec:proximity-priors}
In this section, we model \emph{how far away the next visit will be from the current one}, or the \emph{proximity} distribution. It is the second term of (\ref{eq:two-terms}). First, we approximate that conditional probability with its unconditional counterpart.
\begin{equation}
    \Pr \left[ \mathrm{poi}^{i}_{n_{i}+1} \eqdist{i}{n_i} \mathrm{poi}_{k} \mid \mathbf{v}_i \right] 
    \approx \Pr \left[ \mathrm{poi}^{i}_{n_{i}+1} \eqdist{i}{n_i} \mathrm{poi}_{k} \right] \label{eq:prox-prior}
\end{equation}

To approximate the latter part of (\ref{eq:prox-prior}), we assume that regardless of the user and the POIs, the distance between two consecutive POI visits made by a user roughly follows a certain probability distribution, named \emph{proximity prior} distribution $\mathfrak{P}$:
\begin{equation}
    \mathrm{distance}\left( \mathrm{poi}^{i}_{j}, \mathrm{poi}^{i}_{j+1} \right) \sim \mathfrak{P}\ \ \forall\ i, j
\end{equation}

\begin{figure}[tb]
    \centering
    \includegraphics[width=\linewidth]{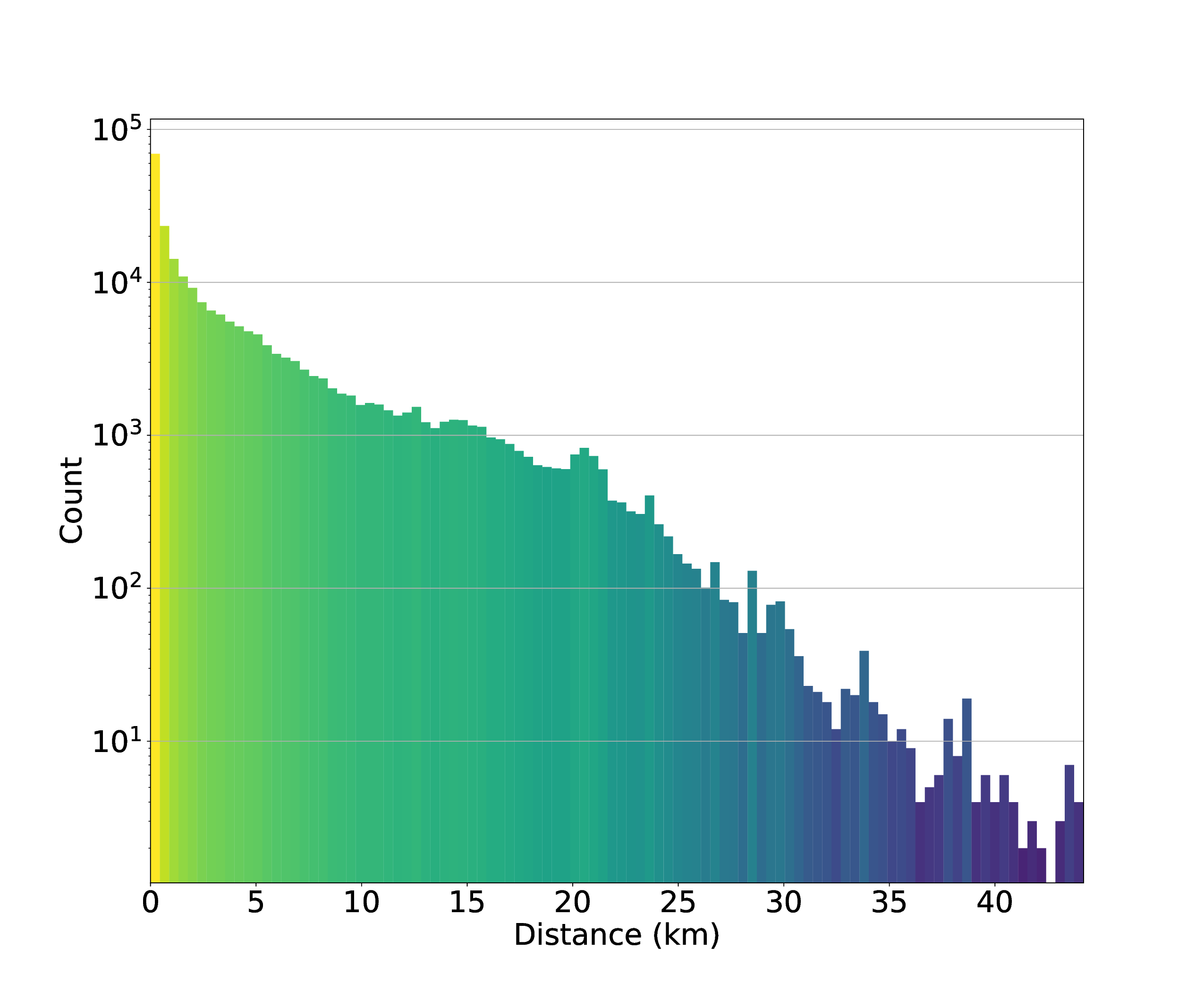}
    \caption{The proximity priors histogram representing the number of instances where the distance between consecutive user locations falls within specific distance intervals (in kilometers).}
    \label{fig:proximity}
\end{figure}

To approximate the continuous proximity prior distribution $\mathfrak{P}$, we produce a discrete probability distribution by grouping the distances consecutive POI visits into counting buckets and normalize the counts to prior probabilities. We plot the proximity priors before normalization in Fig. \ref{fig:proximity}.


Finally, we obtain $\Pr \left[ \mathrm{poi}^{i}_{n_{i}+1} \eqdist{i}{n_i} \mathrm{poi}_{k} \right]$ by combining the approximated proximity prior distribution $\hat{\mathfrak{P}}$, the input visit sequence of the $i$-th user $\mathbf{v}_i$, and the candidate POI $\mathrm{poi}_k$.





\section{Experiments}

\subsection{Experimental Setup}

To evaluate the effectiveness of a POI forecasting model, we follow prior studies to use top-k accuracy, which measure the proportion of ground truth labels appearing in the k classes with the highest probabilities assigned by the method. We denote a top-k accuracy as \textbf{Acc@k}, where $k \in \{1, 5, 10, 20\}$.

\begin{figure}[tb]
    \centering
    \includegraphics[width=\linewidth]{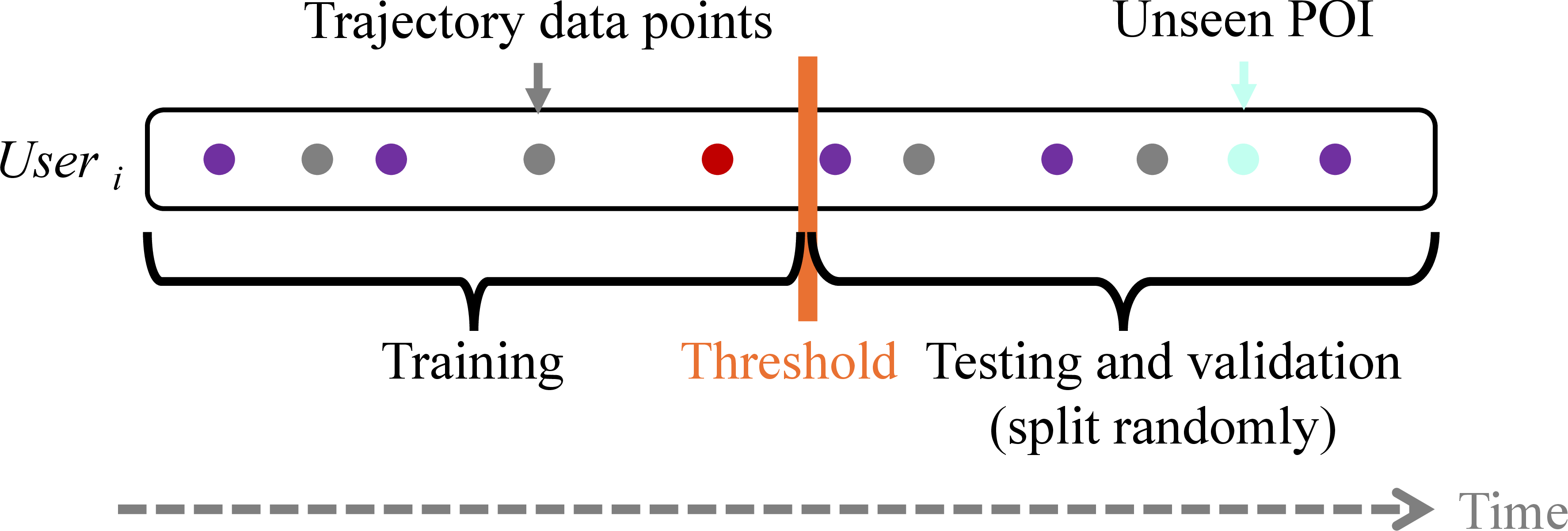}
    \caption{The training, testing, validation, and unseen POI evaluation split.}
    \label{fig:traintestsplit}
\end{figure}

To examine the effect of the ratios between seen and unseen POIs during inference time, we vary them by finding a temporal threshold, all visits before this point were included in the training set. The remaining visits were evenly and randomly split between testing and validation set. POIs that appeared only after the train-test split are considered ``unseen'' and are used to evaluate the ability of an approach to predict new POIs. We illustrate this setup in Fig.~\ref{fig:traintestsplit}.

In our experiments, we used the publicly available, de-identified Foursquare-NYC dataset (FS-NYC)~\cite{yang2014modeling}, which contains 227,428 POI visits in New York city from April 2012 to February 2013.

As for the implementation, we set the length of the observation window to 20, the hidden dimension of Transformer to 128, and the embedding dimensions for POI, semantic, and temporal embeddings to 80, 24, and 24, respectively.
We implement our model in PyTorch and train it with an AMD EPYC 7V13 64-core CPU and an NVIDIA A100 80GB GPU.


\subsection{Results and Discussion}

Table~\ref{tab:results} shows our accuracy in comparison with that of our baseline, MobTCast, when roughly 80\% of POIs are unseen to the models during inference time. Notably, our proposed model achieved 17\% higher top-20 accuracy when evaluating with the set of unseen POIs. Moreover, even if we evaluate both models with all POIs, our method achieves higher accuracy.
This might suggest that our method is more capable of forecasting in a less familiar environment than our baseline.

\begin{table}[tb]
\caption{Top-k accuracy for POI forecasting. The values in column \textbf{POIs} denote the (sub)set of POIs used to produce for \emph{evaluation} in that row.
For all results in this table, we set the training-testing split threshold such that 80\% of all POIs in testing are unseen. The numbers in bold are the best-performing in each POI set and each metric.}\label{tab:results}
\centering
\begin{tabular}{llcccc}\toprule
\textbf{POIs} & \textbf{Method} & \textbf{Acc@1} & \textbf{Acc@5} & \textbf{Acc@10} & \textbf{Acc@20} \\ \midrule
\multirow{2}{*}{All} & MobTCast & 0.0151 & 0.0523 & \textbf{0.1606} & 0.1689 \\ 
                     & Ours     & \textbf{0.0379} & \textbf{0.1130} & 0.1556 & \textbf{0.1986} \\ \midrule
\multirow{2}{*}{Unseen} & MobTCast & 0.0000 & 0.0000 & 0.0000 & 0.0000 \\ 
                        & Ours     & \textbf{0.0333} & \textbf{0.0964} & \textbf{0.1491} & \textbf{0.1789} \\ \bottomrule
\end{tabular}
\end{table}

\begin{figure}[tb]
    \centering
    \includegraphics[width=\linewidth]{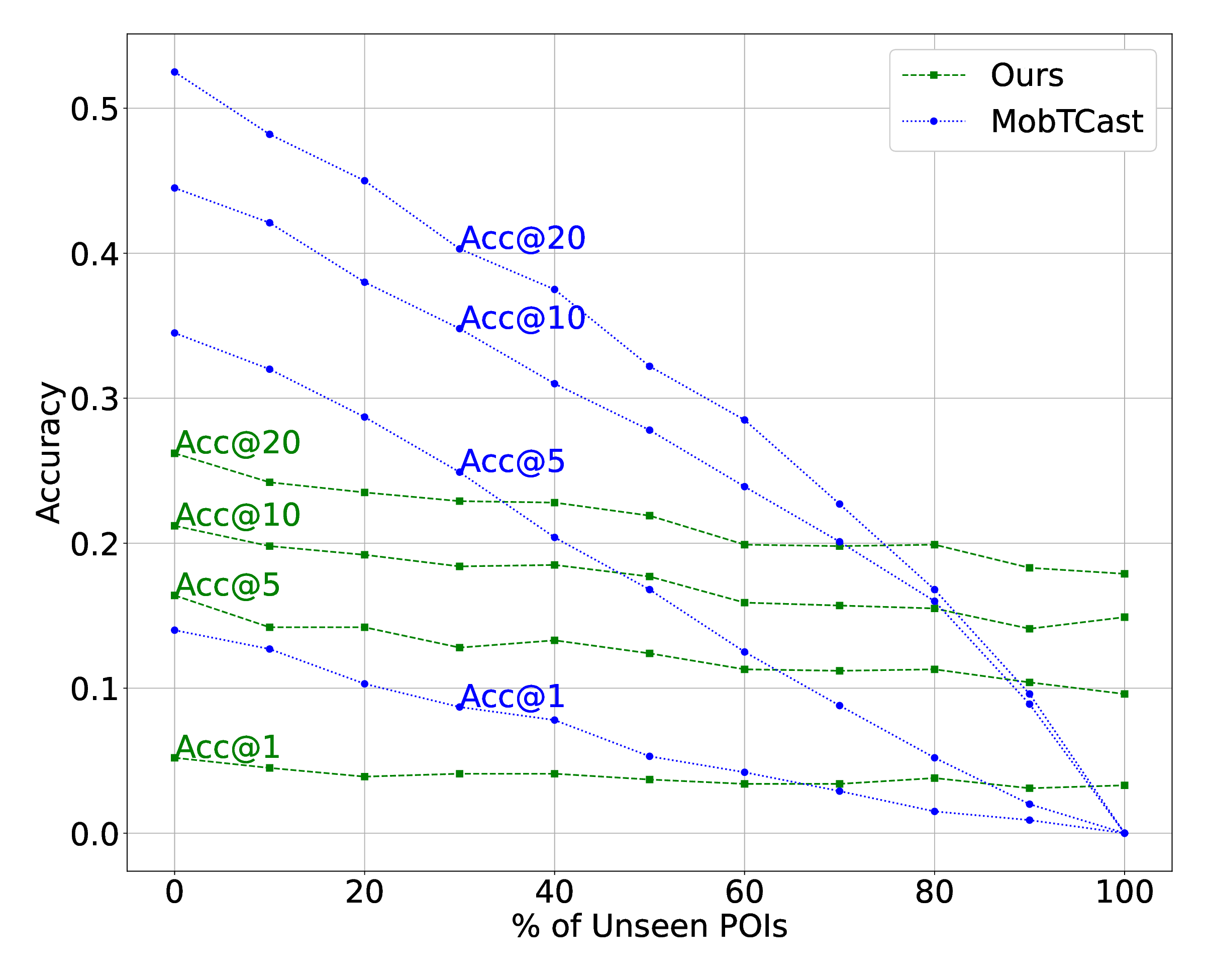}
    \caption{The comparison of MobTCast and our model accuracy as the percentage of unseen POIs increases.\protect\footnotemark}
    \label{fig:percentage}
\end{figure}

We further test the accuracy of different approaches as there are different an increasing number of unseen POIs relative to the total POIs. 
This simulates an environment with a number of emerging new POIs.
As time passes and more unseen POIs has opened, our approach stays more robust compared to our baseline, as demonstrated in Fig. \ref{fig:percentage}.
In fact, regressing these accuracy lines to their line of best fit, the average decrease in slope in 635.417\%. Therefore, our model outperforms tremendously in maintaining accuracy over time as the more unseen POIs appear. 


\footnotetext{The data used for each percentage of unseen POIs are from different train-test split thresholds.}


\subsection{Limitations and Future Work}
The scope of this study is to remain robust with increasing proportion of unseen POIs, which is orthogonal to the studies on the accuracy improvement for those seen POIs. 
We encourage future researchers to explore mechanisms that combines (1) our approach for unseen POIs and (2) another for seen POI, such as MobTCast, to improve the general accuracy for any datasets. For example, one can create a classifier that can dynamically identify whether the next POI will be a seen or unseen one, which could potentially lead to a more comprehensive model for better performance in wider real-world applications.

\section{Conclusion}
Our model, which integrates semantic context prediction and proximity priors, addresses the shortcomings of state-of-the-art models for predicting new POIs unseen during training. 

By combining the semantic context prediction with spatial proximity prediction using a pre-computed proximity priors, our model demonstrates a higher degree of flexibility and adaptability in scenarios where POIs are evolving.

Our results indicate that this approach outperforms the baseline by 17\% in top-20 accuracy when predicting unseen POIs, a critical improvement for real-world applications that has a ever-changing urban landscape. Moreover, our model's accuracy degradation over time, as more unseen POIs are introduced, occurs at a much slower rate compared to existing models. This highlights our model's ability to generalize in the absense of continuously updated trajectory data, making it particularly useful for location-based recommendation systems, as well as as other mobility trajectory prediction applications.

\section*{Acknowledgment}
Thank Maria Despoina Siampou (University of Southern California) for her valuable insights and comments.

\bibliographystyle{IEEEtran}
\bibliography{base,trajgpt}

\end{document}